\documentclass[conference]{IEEEtran}
\IEEEoverridecommandlockouts
\usepackage{cite}
\usepackage{amsmath,amssymb,amsfonts}
\usepackage{algorithmic}
\usepackage{graphicx}
\usepackage{textcomp}
\usepackage{xcolor}
\def\BibTeX{{\rm B\kern-.05em{\sc i\kern-.025em b}\kern-.08em
    T\kern-.1667em\lower.7ex\hbox{E}\kern-.125emX}}

\usepackage{amsfonts}
\usepackage{algorithmic}
\usepackage{algorithm}
\usepackage{array}
\usepackage{subfigure}
\usepackage{textcomp}
\usepackage{stfloats}
\usepackage{url}
\usepackage{verbatim}
\usepackage{color}
\usepackage{multirow}
\usepackage{bbm}
\usepackage{bm}
\usepackage{makecell}
\usepackage{interval}
\usepackage[misc,geometry]{ifsym}
\usepackage{enumitem}

\newlength\savedwidth
\newcommand\whline{\noalign{\global\savedwidth\arrayrulewidth\global\arrayrulewidth 1.0pt}
\hline\noalign{\global\arrayrulewidth\savedwidth}}

\usepackage{colortbl}
\makeatletter
\newcommand*{\rom}[1]{\expandafter\@slowromancap\romannumeral #1@}
\makeatother
\definecolor{tablegray}{gray}{.9}
\definecolor{tableblue}{RGB}{221,238,251}
\definecolor{tablegreen}{RGB}{225,245,245}
\definecolor{baselinecolor}{gray}{.9}

\newcommand{\etal}{\textit{et al}.}
\newcommand{\ie}{\textit{i}.\textit{e}.}
\newcommand{\eg}{\textit{e}.\textit{g}.}

\begin{document}
\title{

An Ensemble Approach to Short-form Video Quality Assessment Using Multimodal LLM
}
\DeclareRobustCommand*{\IEEEauthorrefmark}[1]{%
  \raisebox{0pt}[0pt][0pt]{\textsuperscript{\footnotesize #1}}%
}
\author{
    \IEEEauthorblockN{
        Wen Wen\IEEEauthorrefmark{1}\textsuperscript{\textasteriskcentered}\thanks{\textsuperscript{\textasteriskcentered}Work done during internship at Google.},
        Yilin Wang\IEEEauthorrefmark{2}\textsuperscript{\textrm{\Letter}}\thanks{\textsuperscript{\textrm{\Letter}}Corresponding author.}, 
        Neil Birkbeck\IEEEauthorrefmark{2} and
        Balu Adsumilli\IEEEauthorrefmark{2}
    }
    \IEEEauthorblockA{
        \IEEEauthorrefmark{1}City University of Hong Kong, Hong Kong SAR\\
        \IEEEauthorrefmark{2}Google Inc. Mountain View, CA, USA\\
        % Email: 
            wwen29-c@my.cityu.edu.hk,
            \{yilin, birkbeck, badsumilli\}@google.com
    }
}

\maketitle

\begin{abstract}
The rise of short-form videos, characterized by diverse content, editing styles, and artifacts, poses substantial challenges for learning-based blind video quality assessment (BVQA) models. Multimodal large language models (MLLMs), renowned for their superior generalization capabilities, present a promising solution. This paper focuses on effectively leveraging a pretrained MLLM for short-form video quality assessment, regarding the impacts of pre-processing and response variability, and insights on combining the MLLM with BVQA models. We first investigated how frame pre-processing and sampling techniques influence the MLLM's performance. Then, we introduced a lightweight learning-based ensemble method that adaptively integrates predictions from the MLLM and state-of-the-art BVQA models. Our results demonstrated superior generalization performance with the proposed ensemble approach. Furthermore, the analysis of content-aware ensemble weights highlighted that some video characteristics are not fully represented by existing BVQA models, revealing potential directions to improve BVQA models further.
\end{abstract}

\begin{IEEEkeywords}
Video quality assessment, short-form video, multimodal large
language model, content-aware ensemble
\end{IEEEkeywords}

\section{Introduction}~\label{section:intro}
Driven by advancements in technology and the high popularity of social media, the increasing accessibility of video creation and editing tools has led to a surge in user-generated short-form videos. 
Short-form videos are designed for quick consumption and sharing on social media platforms such as YouTube Shorts, TikTok, and Instagram Reels.
These videos exhibit a wide range of content, compositions, and editing styles. While they enhance viewer engagement, their diversity presents substantial challenges for blind video quality assessment (BVQA) research.

\begin{figure}%[!tbp]
% \scriptsize
\setlength{\abovecaptionskip}{0cm}
\setlength{\belowcaptionskip}{-0.35cm}
\centering
\includegraphics[width=0.45\textwidth]{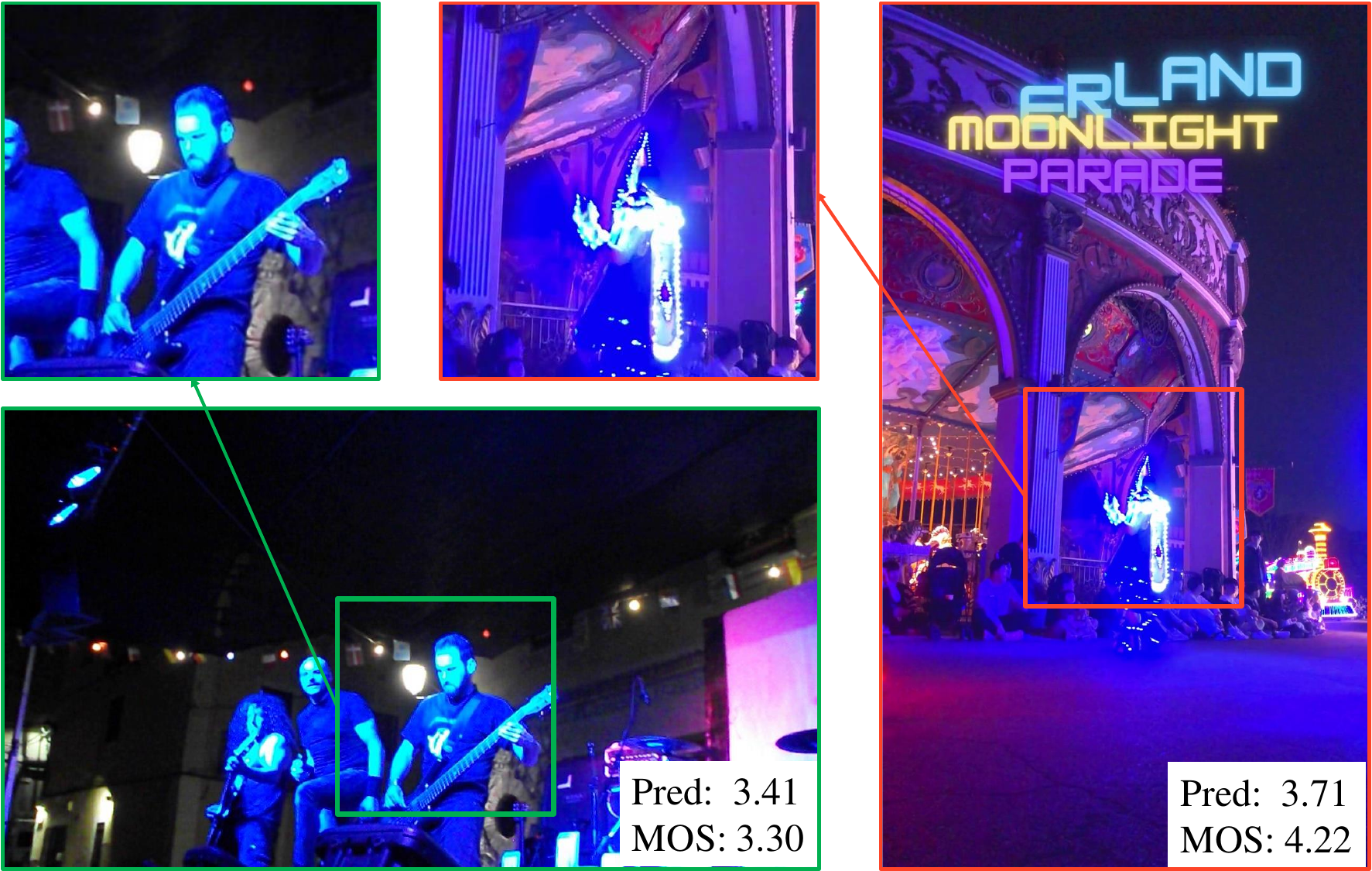}
\caption{The video in the green frame is from YouTube-UGC~\cite{wang2019youtube}, while the video in the red frame is sourced from Shorts-SDR~\cite{wang2024youtube}. Despite sharing similar local artifacts, the short-form video contains more edited content and demonstrates a significantly higher mean opinion score (MOS). However, the learning-based model generates similar predictions for both videos, ultimately underestimating the quality of the short-form video.
}
\vspace{-12pt}
\label{fig:shorts_compare}
\end{figure}

BVQA models encompass both knowledge-based models~\cite{saad2014blind, mittal2015completely, korhonen2019two, tu2021ugc} and learning-based models~\cite{li2019quality, li2022blindly, ying2021patch, wang2021rich, sun2022deep, wu2022fast, wu2022disentangling, sun2024analysis, wen2024modular}, showcasing substantial progress. 
However, a recent study~\cite{wang2024youtube} highlighted that even the cutting-edge learning-based models encounter difficulties adapting to newly introduced short-form datasets. 
Figure~\ref{fig:shorts_compare} illustrates this discrepancy, contrasting a YouTube-UGC~\cite{wang2019youtube} video (green frame) with a Shorts-SDR~\cite{wang2024youtube} video (red frame). 
Despite these two videos sharing similar local artifacts, the short-form video contains more edited content, prompting viewers to tolerate artifacts more and resulting in a significantly higher mean opinion score (MOS). 
However, when a learning-based model such as FasterVQA~\cite{wu2023neighbourhood} is applied, it generates similar predictions for both videos, thereby underestimating the quality of the short-form video. 
This finding implies that BVQA models may encounter challenges in generalization due to their limited understanding of evolving video content.

Multimodal large language models (MLLMs) have recently experienced explosive growth, demonstrating enhanced generalization capabilities with better content understanding and proficiency in various multitasking domains~\cite{zhang2023internlm, ye2024mplug, dong2024internlm, liu2024visual, peng2024grounding, zhu2024minigpt}. 
Several studies have concentrated on benchmarking or proposing MLLM models for quality assessment. For example, Wu~\etal~\cite{wu2024qbench}  assessed the quality assessment capabilities of multiple MLLMs by creating a novel dataset with human descriptions. In another study, Wu~\etal~\cite{wu2024comprehensive} benchmarked the quality assessment abilities of multiple MLLMs by altering the prompting strategies. Recent research efforts\cite{you2023depicting, wu2024qinstruct, wu2024qalign, wu2024towards, liu2024improved} have demonstrated that instruction tuning with human-annotated datasets can significantly enhance the performance of publicly available MLLMs in quality assessment. Furthermore, a recent study by Ge~\etal~\cite{ge2024lmm} introduced a model that merges spatial and temporal features from  BVQA models with embeddings from MLLMs to achieve superior performance.

However, certain areas remain unexplored. For example, the quality of images and videos is closely related to spatial pre-processing~\cite{mackin2018studyreso, li2019avc}, yet limited research explores how MLLMs handle input data. 
Additionally, previous studies~\cite{wu2024qbench, wu2024comprehensive} only elicited responses from MLLMs once, neglecting the potential for more robust outcomes through repeated evaluations. 
Moreover, how MLLMs perceive short-form video quality and its potential to uncover deficiencies in existing models are also worth investigating.
% In summary, 
This study aims to address these inquiries and presents the following contributions:
\begin{itemize}[leftmargin=*]
\item We explored the influence of frame pre-processing and robust inference techniques on a comparatively lightweight MLLM regarding short-form video quality assessment;
\item We proposed a content-aware ensemble mechanism to integrate predictions from the MLLM and learning-based models. Our results demonstrate superior generalization performance in short-form datasets;
% in compared to using either approach independently;
\item We identified specific video samples where our approach is most beneficial by analyzing the learned weights of our ensemble method. This analysis offers insights into how an MLLM can compensate for the limitations of existing learning-based BVQA models.
\end{itemize}

\begin{figure*}%[!tbp]
% \scriptsize
\setlength{\abovecaptionskip}{0cm}
\setlength{\belowcaptionskip}{-0.35cm}
\centering
\includegraphics[width=0.90\textwidth]{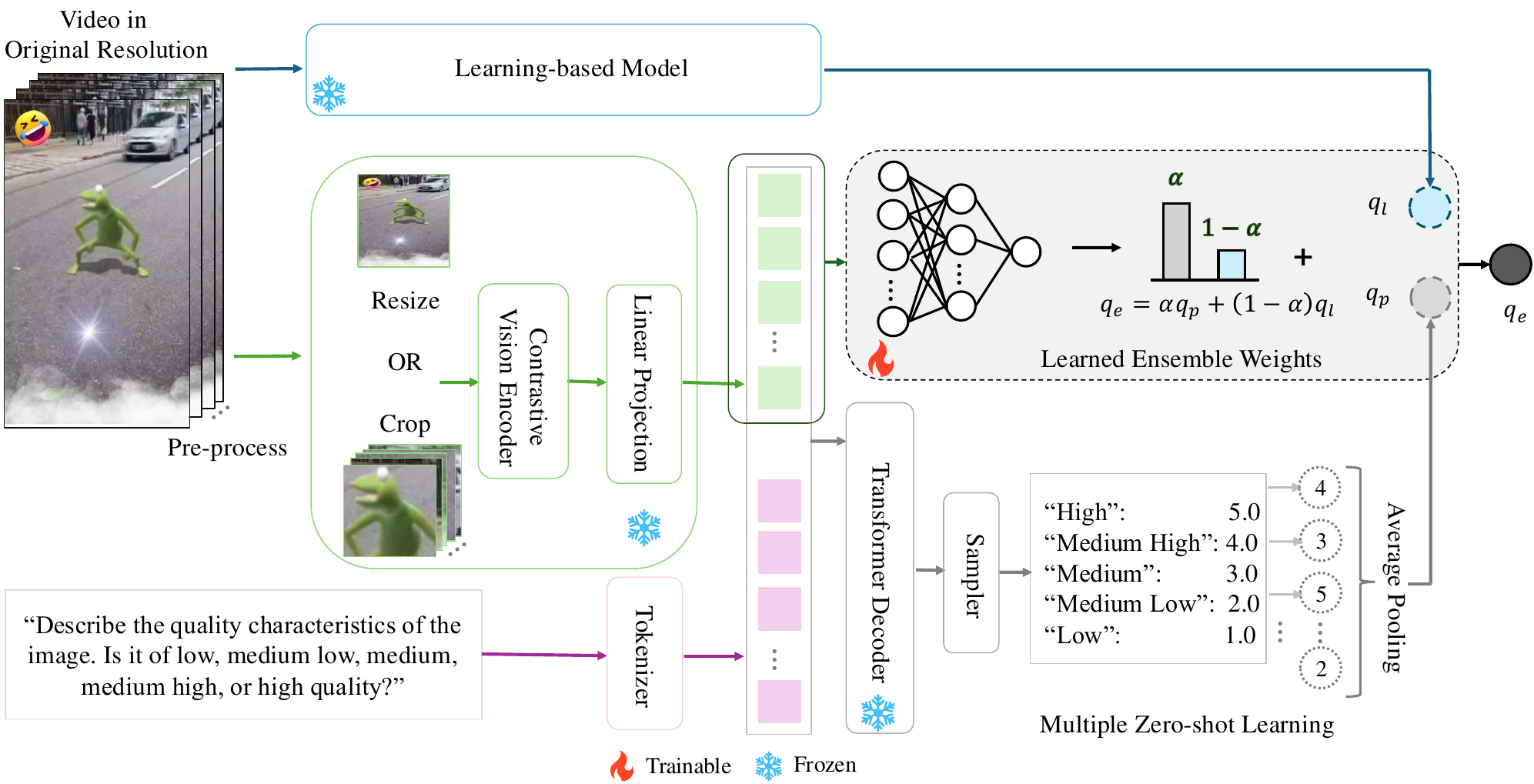}
\caption{The system diagram illustrates our MLLM evaluation methodology and the content-aware ensemble method, with only the gray part being tuned.
In the MLLM evaluation process, videos are initially downsampled to key frames, which serve as input. These key frames are subsequently resized or cropped and subjected to multiple zero-shot prompting experiments employing various sampler techniques. The resulting numerous outputs are aggregated to produce a score, with the average serving as the MLLM's final prediction $q_p$.
In the content-aware ensemble method, only the image features from the vision encoder are employed. A learned weight $\alpha$ is calculated to ensemble the predictions $q_p$ from the MLLM with the predictions $q_l$ from the learning-based models.
}
\label{fig:mllm_prompt}
\vspace{-12pt}
\end{figure*}

\section{MLLM Evaluation Approach}\label{section:mllm_eval}
As highlighted in Section~\ref{section:intro}, prior studies have typically conducted a single inference with MLLMs using score-related prompts only. In this section, we explored the potential impacts of prompting, image pre-processing, robust inference strategies, and the number of zero-shot trials on the MLLM's output in quality assessment. 
The methodology and results are outlined below.

\subsection{Methodology}

\noindent\textbf{Dataset}. 
We employed the most recent short-form video quality datasets, Shorts-SDR and Shorts-HDR2SDR~\cite{wang2024youtube}. These two datasets are exclusively sourced from YouTube Shorts, which are distinct from public video databases. While alternative short-form video datasets~\cite{zhang2023md, lu2024kvq} primarily focus on live broadcast distortions, compression artifacts, and enhancements, Shorts-SDR and Shorts-HDR2SDR concentrate on a broader spectrum of user-generated content, providing the most enormous short-form video datasets currently available.

\noindent\textbf{MLLM}. 
To enable swift consecutive inferences, we assessed PaliGemma~\cite{beyer2024paligemma}, a recently developed MLLM acclaimed for its state-of-the-art performance across diverse tasks, despite its relatively modest scale of $3$ billion parameters. 
Instead of connecting frozen image components and language components with lightweight adapters~\cite{jaegle2021perceiver, alayrac2022flamingo, li2023blip}, PaliGemma trained both the vision component SigLIP~\cite{chen2023pali} and language component Gemma~\cite{team2024gemma} jointly, and functioned as a cohesive unit. 
Among the several provided checkpoints, we select the one trained on a mix of tasks with an input resolution of $448^2$, purportedly more effective for tasks demanding higher-resolution input.

\noindent\textbf{Pre-processing strategy}. 
Given that the input size of PaliGemma is $448^2$, and the videos in Shorts-SDR and Shorts-HDR2SDR are all $1080$P, frame pre-processing (\eg resizing or cropping) is required to fit the model input size. Resizing retains all semantic content but may change the aspect ratio while cropping preserves more native details but only from local regions. Both approaches were investigated in this section. 

\noindent\textbf{Robust inference technique}. 
For robust inference, it is essential to customize samplers with randomization factors for multiple zero-shot scenarios. A sampler refers to an algorithm that governs how an MLLM chooses the next token based on the probabilities the model provides. Common sampler variations include greedy, temperature, and nucleus. While the greedy sampler picks tokens with the highest probabilities, temperature and nucleus samplers, with increased randomness factors, often yield more creative responses. Here, we enabled nucleus samplers in model inference due to their capacity to generate diverse responses through multiple inferences. 

\begin{table}%\footnotesize
\setlength{\abovecaptionskip}{0.0cm}
\setlength{\belowcaptionskip}{0.15cm}
\caption{Performance comparison of different prompting, pre-processing \& sampler strategies. \rom{1} \& \rom{2} represent score-related prompt \& level related prompt respectively. All numbers are presented in the SRCC / PLCC format. The top-$1$ results on each dataset are highlighted in \textbf{bold}.}
\label{tab:process_sampler}
\centering
\renewcommand{\arraystretch}{1.25}
% \resizebox{1\textwidth}{!}{
% \tabcolsep=3pt
\begin{tabular}{lccc}
% \toprule[.15em]
% \whline
Pre-processing \& Sampler & Shorts-SDR & Shorts-HDR2SDR\\
\whline
% Resize-Score-Nucleus$(0.0)$  &  &  \\
Resize-Nucleus$(0.0)$-\rom{1}  & 0.252 / 0.082 &  0.045 / 0.024 \\
Resize-Nucleus$(0.0)$-\rom{2}  & 0.375 / 0.438 &  0.299 / 0.323 \\
Resize-Nucleus$(0.2)$-\rom{2}  & 0.342 / 0.395 & 0.299 / 0.323  \\ 
Resize-Nucleus$(0.5)$-\rom{2}   &  0.361 / 0.425 & 0.310 / 0.339  \\ 
Resize-Nucleus$(0.9)$-\rom{2}  & 0.375 / 0.456 &  0.318 / 0.347 \\
% Resize-Nucleus$(0.2)$  & 0.342 / 0.395 & 0.299 / 0.323  \\ 
% Resize-Nucleus$(0.5)$   &  0.361 / 0.425 & 0.310 / 0.339  \\ 
% Resize-Nucleus$(0.9)$  & 0.375 / 0.456 &  0.318 / 0.347 \\
% Resize-Temperature$(0.2)$  & 0.360 / 0.413 & 0.308 / 0.330  \\
% Resize-Temperature$(0.5)$  &  0.372 / 0.434 & 0.314 / 0.340  \\
% Resize-Temperature$(0.9)$  & 0.374 / 0.447 & 0.318 / 0.344  \\
% \hline
Crop-Nucleus$(0.0)$-\rom{2} & 0.599 / 0.567 &  0.522 / 0.546 \\
Crop-Nucleus$(0.2)$-\rom{2}  & 0.626 / 0.593 & 0.524 / 0.561 \\
Crop-Nucleus$(0.5)$-\rom{2}  & 0.635 / 0.609 & 0.532 / 0.577  \\
Crop-Nucleus$(0.9)$-\rom{2}  &  \textbf{0.677} / \textbf{0.679} & \textbf{0.534} / \textbf{0.601}  \\
% Crop-Nucleus$(0.2)$  & 0.626 / 0.593 & 0.524 / 0.561 \\
% Crop-Nucleus$(0.5)$  & 0.635 / 0.609 & 0.532 / 0.577  \\
% Crop-Nucleus$(0.9)$  &  \textbf{0.677} / \textbf{0.679} & \textbf{0.534} / \textbf{0.601}  \\
% Crop-Temperature$(0.2)$ & 0.630 / 0.599 & 0.527 / 0.569  \\
% Crop-Temperature$(0.5)$  & 0.659 / 0.647 & 0.532 / 0.600 \\
% Crop-Temperature$(0.9)$  & 0.665 / 0.670 & 0.533 / 0.601  \\

% \whline

\end{tabular}
% }
\vspace{-12pt}
\end{table}

\subsection{MLLM Zero-shot Experiment \& Analysis}
We first extracted frames from each video at a rate of one frame per second, resulting in $5$ key frames for videos in the the two short-form datasets. 
Concerning prompting, the first prompt is score-related: \textit{``Describe the quality characteristics of the image. Rate the image quality on a scale of 1 to 5, with 1 being the lowest quality and 5 being the highest quality.}''. 
We compared it with a level-related prompt: \textit{``Describe the quality characteristics of the image. Is it of low, medium low, medium, medium high, or high quality?}''.
For resizing, we downscaled each key frame to size $448^2$ using bilinear interpolation without keeping the aspect ratio.
Regarding cropping, we randomly segmented each key frame into $10$ patches of size $448^2$.
Each resized image or cropped patch undergoes $20$ rounds of zero-shot inferences on the MLLM. 
In terms of sampler strategies, we employ a nucleus sampler of $0.0$, $0.2$, $0.5$, and $0.9$.
Upon receiving responses, we filtered out nonsensical answers and assigned a score ranging from $1$ to $5$ to each level for the second prompt. 
% with higher scores denoting superior quality. 
Subsequently, we computed the average of these scores from each video, serving as PaliGemma's final prediction for the video. 
To evaluate the randomness of predictions, we varied the number of zero-shot trials in each frame to observe the standard deviations.

\begin{figure}%[!tbp]
% \scriptsize
\setlength{\abovecaptionskip}{0cm}
\setlength{\belowcaptionskip}{-0.35cm}
\centering
\includegraphics[width=0.42\textwidth]{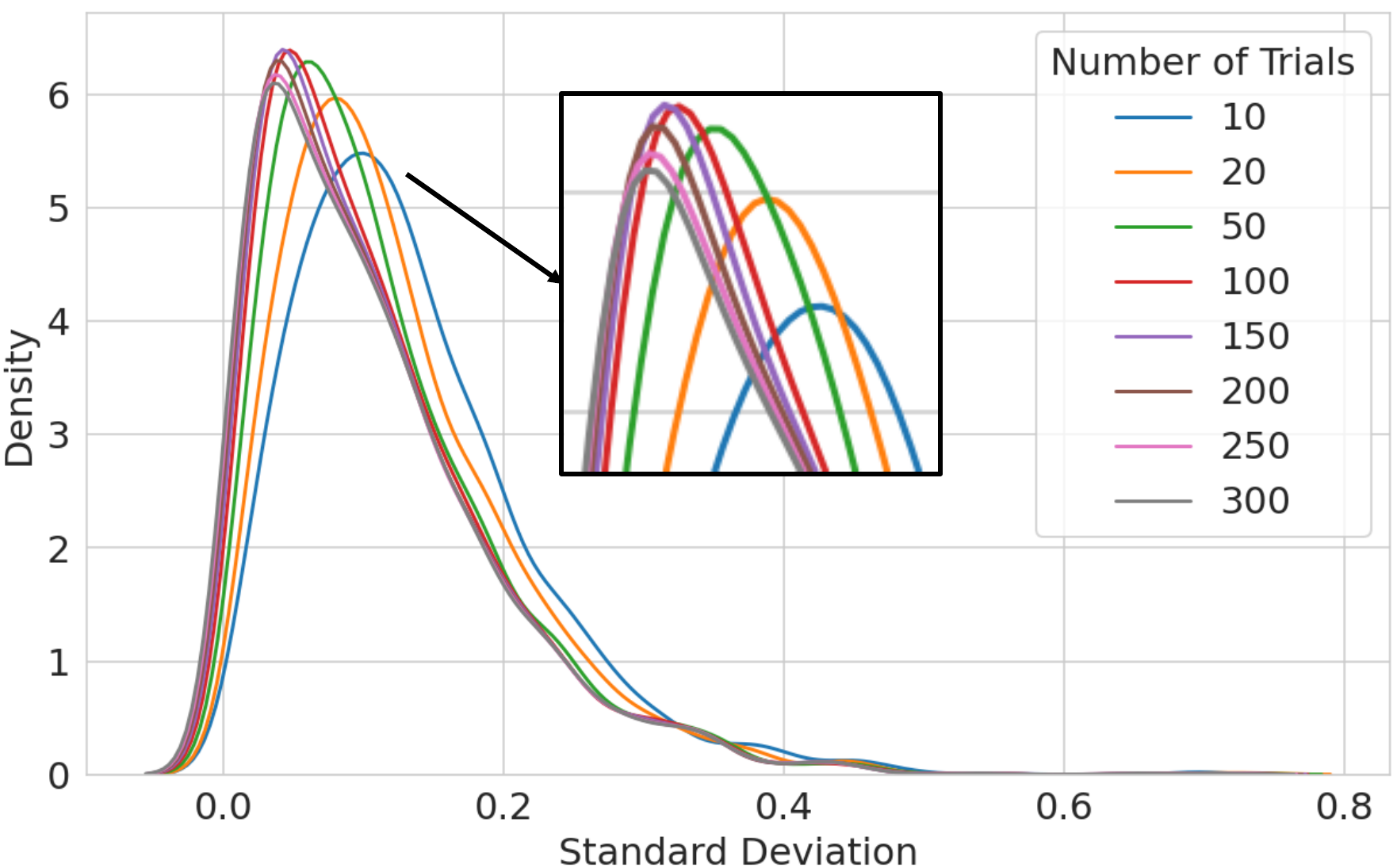}
\caption{Distributions of standard deviations across varying numbers of zero-shot trials per frame.  
Level-related prompting, cropping pre-processing, and a nucleus sampler set at $0.9$ are utilized. The default trial number is $200$, involving $10$ crops with $20$ trials each.
}
\vspace{-12pt}
\label{fig:repeat}
\end{figure}

% \subsection{Analysis}
The experiment results are presented in Table~\ref{tab:process_sampler}. The distribution of standard deviations is depicted in Figure~\ref{fig:repeat}. By examining these results, we arrived at some observations.
\begin{itemize}[leftmargin=*]
\item The level-related prompt yields significantly superior results compared to the score-related prompt, indicating that MLLM can discern quality with the level-related prompt. 
The restricted capacity to provide precise numerical values with score-related prompts hinders MLLM's overall performance in quality assessment.
\item Cropping exhibits significantly superior outcomes compared to resizing. However, few quality studies have utilized cropped patches as inputs for MLLM.
\item Enhanced outcomes are observed with higher randomness factors, albeit with marginal improvements. 
\item Given that the scores range from $1$ to $5$, the minor standard deviations in Figure~\ref{fig:repeat} indicate the stability of MLLM outputs during multiple zero-shot trials. Moreover, as the number of zero-shot trials per frame increased, narrower standard deviation distributions were observed, underscoring the importance of numerous zero-shot trials in ensuring robust predictions.
\end{itemize}

\section{Content-aware Ensemble Framework}
MLLM exhibited promising correlations with MOS. Building upon this finding, we want to know if MLLM simply replicates existing BVQA models, or offers insights not captured by these models.
To this end, we designed a content-aware ensemble framework to investigate the relation between the predictions from MLLM and existing models.

\subsection{A Content-aware Ensemble Approach}

We define a video sequence as $\bm{x}= \{\bm x_i\}_{i=0}^{N-1}$, where $\bm x_i\in\mathbb{R}^{H\times W \times 3}$ represents the $i$-th frame, with $H$ and $W$ denoting the frame height and width, and $N$ representing the total number of frames.
The content-aware ensemble module $f(\cdot):\mathbb{R}^{H\times W\times 3\times N}\mapsto \mathbb{R}$ takes $\bm{x}$ as input and generates a weight $\alpha$ within the range of [0,1]. The function $f(\cdot)$ primarily captures semantic visual content not influenced by spatial resizing.
To be specific, we uniformly sample a sparse set of $M$ key frames, $\bm y = \{\bm y_i\}_{i=0}^{M-1}$, which undergo bilinear resizing to $H_b\times W_b$ without preserving the aspect ratio. Here, $H_b$ and $W_b$ are determined by the image input of  SigLIP~\cite{chen2023pali}, the visual encoder of PaliGemma. Additionally, we integrate a multilayer perceptron (MLP) comprising two layers (\ie, two fully connected layers with ReLU nonlinearity in between) and a sigmoid layer to ensure that the weight $\alpha$ falls within the appropriate range. The features extracted by SigLIP from $M$ key frames are concatenated, spatially pooled, and fed to the two-layer MLP to produce a content-aware weight $\alpha$.

Utilizing the learned weight, we can directly apply the weighted combination of the MLLM's predictions and predictions from existing BVQA models:
\begin{align}\label{eq:ensemble}
q_e = \alpha q_p + (1-\alpha) q_l ,
\end{align}
where $q_p$ is the prediction from PaliGemma, $q_l$ is the prediction from a learning-based model, $q_e$ represents the ensemble score.

\begin{table}%\small%\footnotesize%\scriptsize
\setlength{\abovecaptionskip}{0.0cm}
\setlength{\belowcaptionskip}{0.15cm}
\centering
\renewcommand{\arraystretch}{1.25}
\caption{Inference performance comparison of our content-aware ensemble method against four competing models on Shorts-SDR and  Shorts-HDR2SDR. Na\"ive \& proposed means na\"ive \& the proposed ensemble method respectively.
}
\label{tab:performance_ensemble}
% \resizebox{1\textwidth}{!}{
\begin{tabular}{lcc}
% \whline
Method   & Shorts-SDR & Shorts-HDR2SDR  \\
\whline

FastVQA~\cite{wu2022fast}   & 0.793 / 0.802 &  0.530 / 0.652  \\ 
FastVQA-Na\"ive   & 0.802 / 0.815 & 0.560 / 0.674 \\ 
FastVQA-Proposed   &  \textbf{0.808} / 0.828  & 0.582 / 0.679 \\ 
\hline
FasterVQA~\cite{wu2023neighbourhood}  &  0.756 / 0.783 &  0.494 / 0.601 \\ 
FasterVQA-Na\"ive     & 0.782 / 0.808 &  0.538 / 0.660 \\ 
FasterVQA-Proposed    &  0.791 / 0.825 & 0.569 / 0.663  \\ 
\hline
DOVER~\cite{wu2022disentangling}    & 0.768 / 0.801 &  0.512 / 0.614 \\ 
DOVER-Na\"ive    & 0.799 / 0.831 & 0.551 / 0.682 \\ 
DOVER-Proposed    &  \textbf{0.808} / 0.839 & 0.584 / 0.683 \\ 
\hline
ModularBVQA~\cite{wen2024modular}   &  0.753 / 0.807 &  0.559 / 0.686 \\ 
ModularBVQA-Na\"ive   & 0.782 / 0.810 & 0.592 / 0.711 \\ 
ModularBVQA-Proposed    & 0.807 / \textbf{0.847} & \textbf{0.612} / \textbf{0.723} \\ 

% \bottomrule[.15em]
% \whline
\end{tabular}
% }
\vspace{-12pt}
\end{table}

\subsection{Ensemble Experiment \& Analysis}
\noindent\textbf{Learning-based Models}. 
We evaluated four prominent BVQA models: FastVQA~\cite{wu2022fast}, FasterVQA~\cite{wu2023neighbourhood}, DOVER~\cite{wu2022disentangling}, and ModularBVQA~\cite{wen2024modular}. Each model was initialized with checkpoints pretrained on LSVQ~\cite{ying2021patch}, provided by the original authors. We conducted inference experiments using these models without fine-tuning them on short-form videos.

\noindent\textbf{Implementation Details}. 
We initially performed a na\"ive combination with $\alpha$ set to $0.5$ for reference.
For the content-aware ensemble approach, adhering strictly to an inference-only setup, we exclusively tuned the ensemble module on LSVQ, as shown in Figure~\ref{fig:mllm_prompt}. The number of key frames, $M$, is set to $5$.
All key frames underwent bilinear resizing to dimensions of $448^2$. Optimization of all trainable parameters was carried out using the Adam method across $10$ epochs, commencing with an initial learning rate of $3 \times 10^{-4}$ and decay at a rate of $0.95$ every $2$ epochs, with a minibatch size of $32$. We employed $\ell_2$ as the optimization goal. Upon completion of training, we utilized the checkpoint for evaluation on the two short-form datasets to generate the ensemble score.

% \subsection{Analysis}
The results for the BVQA models only, na\"ive combination, and our proposed method are presented in Table~\ref{tab:performance_ensemble}. We further identify specific instances from Shorts-SDR and Shorts-HDR2SDR where the learned weight $\alpha$ for MLLM's prediction $q_p$ is significant, accompanied by a considerable absolute difference $|q_p-q_l|$. 
We choose $q_l$ from FasterVQA, where our ensemble method shows the most significant enhancement. 
Examination of these results provides valuable insights.

\begin{figure}%[!tbp]
% \scriptsize
\setlength{\abovecaptionskip}{0cm}
\setlength{\belowcaptionskip}{-0.35cm}
\centering
\includegraphics[width=0.40\textwidth]{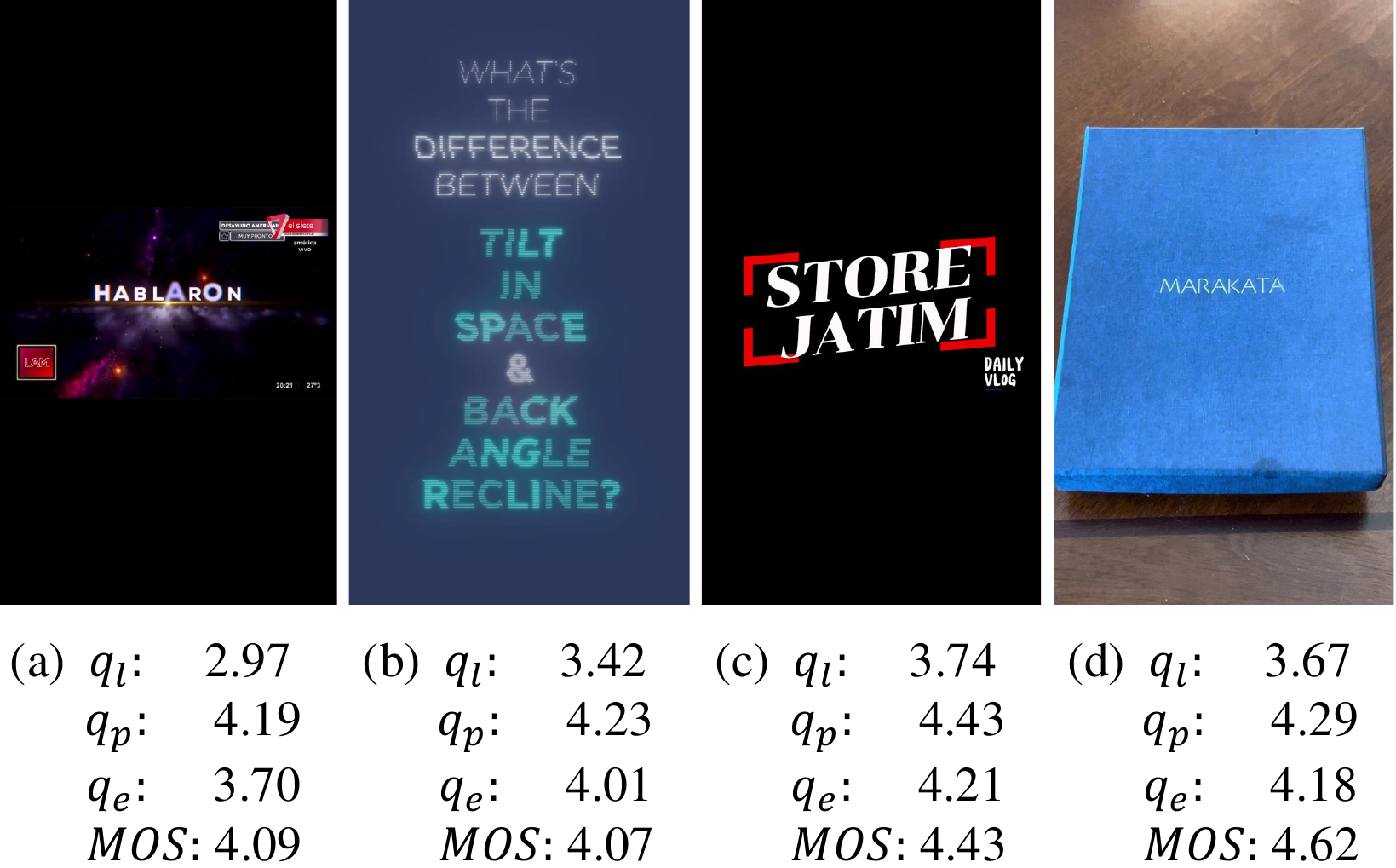}
\caption{Thumbnails of videos where BVQA models tend to \textbf{underestimate}, yet with MLLM can provide superior predictions. (a) and (b) are from Shorts-SDR, while (c) and (d) are from Shorts-HDR2SDR.}
\vspace{-3pt}
\label{fig:shorts_under}
\end{figure}

\begin{figure}%[!tbp]
% \scriptsize
\setlength{\abovecaptionskip}{0cm}
\setlength{\belowcaptionskip}{-0.35cm}
\centering
\includegraphics[width=0.40\textwidth]{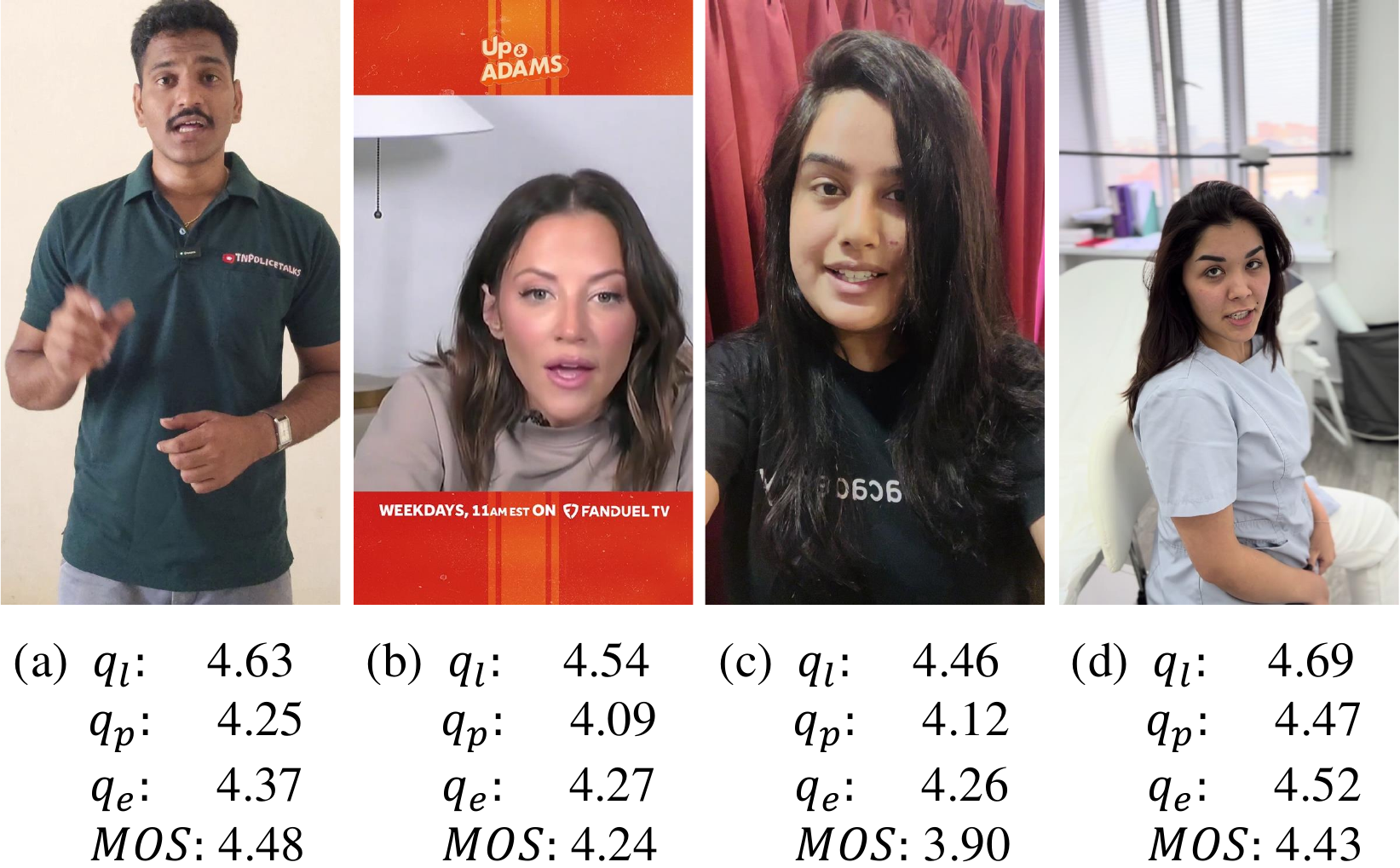}
\caption{Thumbnails of videos where BVQA models tend to \textbf{overestimate}, yet with MLLM can provide superior predictions. (a) and (b) are from Shorts-SDR, while (c) and (d) are from Shorts-HDR2SDR.}
\vspace{-12pt}
\label{fig:shorts_over}
\end{figure}

\begin{itemize}[leftmargin=*]
\item The learning-based models demonstrate satisfactory performance in Shorts-SDR; however, they exhibit mediocre results in Shorts-HDR2SDR, 
% comparable to or 
even inferior to MLLM. 
\item Even a na\"ive combination can yield a performance that surpasses either the learning-based model or the MLLM individually, underscoring the significance of MLLM coordinating with learning-based models.
\item The performance of our proposed ensemble method consistently outperforms a na\"ive combination, highlighting the superiority of our content-aware ensemble approach.
\item BVQA models tend to underestimate videos containing animation-like content, special effects, or scenes with predominantly dark areas. In contrast, MLLM can 
% recognize such content and 
give higher predictions, as illustrated in Figure~\ref{fig:shorts_under}.
\item BVQA models tend to overestimate videos where humans are the primary subjects positioned at the center. 
% of the screen. 
Conversely, MLLM demonstrates greater resilience to such content and offers lower predictions, as shown in Figure~\ref{fig:shorts_over}.
\end{itemize}

\section{Conclusion}
In this study, we harnessed an MLLM's capabilities for assessing short-form video quality by refining pre-processing techniques and sampling strategies. Additionally, we introduced a content-aware ensemble approach to combine predictions from the MLLM and learning-based models.  
Despite its simplicity, the learned weights provided valuable insights into existing models' challenges in generalizing effectively to novel content in short-form videos. 
Notably, the MLLM displayed increased robustness to such unique content, thereby assisting BVQA models in generating enhanced performance.

% \section*{Acknowledgment}

% The preferred spelling of the word ``acknowledgment'' in America is without 
% an ``e'' after the ``g''. Avoid the stilted expression ``one of us (R. B. 
% G.) thanks $\ldots$''. Instead, try ``R. B. G. thanks$\ldots$''. Put sponsor 
% acknowledgments in the unnumbered footnote on the first page.

\vfill\pagebreak

% References should be produced using the bibtex program from suitable
% BiBTeX files (here: strings, refs, manuals). The IEEEbib.bst bibliography
% style file from IEEE produces unsorted bibliography list.
% -------------------------------------------------------------------------

\bibliographystyle{IEEEtran}
\bibliography{refs}{}

\end{document}